\documentclass[conference]{IEEEtran}
\IEEEoverridecommandlockouts
\usepackage{cite}
\usepackage{amsmath,amssymb,amsfonts}
\usepackage{graphicx}
\usepackage{textcomp}
\usepackage{xcolor}
\usepackage{cleveref}

\usepackage{algorithm}
\usepackage[noend]{algpseudocode}
\def\BibTeX{{\rm B\kern-.05em{\sc i\kern-.025em b}\kern-.08em
    T\kern-.1667em\lower.7ex\hbox{E}\kern-.125emX}}
\begin{document}

\title{Automatic Navigation and Voice Cloning Technology Deployment on a Humanoid Robot
\thanks{Identify applicable funding agency here. If none, delete this.}
}

 \author{\IEEEauthorblockN{1\textsuperscript{st} Dongkun Han*}
 \IEEEauthorblockA{\textit{Department of Mechanical and Automation Engineering} \\
 \textit{The Chinese University of Hong Kong}\\
 Hong Kong SAR, China \\
 dkhan@mae.cuhk.edu.hk}
 *Corresponding author
 ~\\
 \and
 \IEEEauthorblockN{2\textsuperscript{nd} Boyuan Shao}
 \IEEEauthorblockA{\textit{Department of Mechanical and Automation Engineering} \\
 \textit{The Chinese University of Hong Kong}\\
 Hong Kong SAR, China \\
 b144792@cuhk.edu.hk}
 }

\maketitle

\begin{abstract}
Mobile robots have shown immense potential and are expected to be widely used in the service industry. The importance of automatic navigation and voice cloning cannot be overstated as they enable functional robots to provide high-quality services. The objective of this work is to develop a control algorithm for the automatic navigation of a humanoid mobile robot called Cruzr, which is a service robot manufactured by Ubtech. Initially, a virtual environment is constructed in the simulation software Gazebo using Simultaneous Localization And Mapping (SLAM), and global path planning is carried out by means of local path tracking. The two-wheel differential chassis kinematics model is employed to ensure autonomous dynamic obstacle avoidance for the robot chassis. Furthermore, the mapping and trajectory generation algorithms developed in the simulation environment are successfully implemented on the real robot Cruzr. The performance of automatic navigation is compared between the Dynamic Window Approach (DWA) and Model Predictive Control (MPC) algorithms. Additionally, a mobile application for voice cloning is created based on a Hidden Markov Model, and the proposed Chatbot is also tested and deployed on Cruzr. 
\end{abstract}

\begin{IEEEkeywords}
Automatic navigation, Voice cloning, Humanoid mobile robot, Chatbot development
\end{IEEEkeywords}

\section{Introduction}
Mobile chassis robots have become increasingly prevalent in the modern service industry, particularly in indoor environments, like hotels, classrooms and restaurants. This is mainly attributed to their affordability and adaptability to various flat surfaces. Typically, a fusion perception scheme that combines vision and radar sensors is employed to generate accurate mapping and positioning information. However, due to the high cost and computational requirements of visual data processing, there are instances where a pure Light Detection And Ranging (LiDAR) solution can be just as effective in indoor environments. These environments often have controlled settings that allow for the creation of a comprehensive room map. Choosing a LiDAR-based solution can be advantageous in terms of cost and computing power limitations. However, mapping, motion planning, and control could be complicated thus more challenging when using LiDAR navigation systems. To confront of this issue, more efforts have been devoted in this area.

When performing Simultaneous Localization and Mapping (SLAM), robots typically utilize motion information from their kinematics or dynamics model along with data acquired from onboard sensors to achieve precise positioning and map construction. SLAM technology finds applications not only in robotics but also in fields like autonomous driving and drones. SLAM can be classified into vision-based schemes, which rely on binocular cameras, monocular cameras, or depth RGB cameras, and LiDAR-based schemes. One approach in vision-based schemes is to construct sparse matrices to obtain matching feature points in different image frames, as seen in Parallel Tracking And Mapping (PTAM) and real-time monocular SLAM. Another approach utilizes image brightness information to build maps using intensive methods \cite{Taketomi2017Visual}. Undoubtedly, vision-based SLAM schemes offer numerous advantages. The camera requirements are not overly demanding, allowing the use of cost-effective cameras, thereby reducing expenses. Additionally, the regular pixel distribution in images enables the utilization of data structures to store information, facilitating flexible map construction through optimized methods. However, compared to vision-based solutions, LiDAR-based solutions tend to provide higher accuracy and better performance in scenarios requiring high-speed movement, such as autonomous driving and drones. LiDAR schemes can be categorized based on the data types provided by the radar, mainly involving two-dimensional or three-dimensional point cloud data. LiDAR captures point cloud data in each frame and employs iterative nearest point (ICP) and normal distribution transform (NDT) to match corresponding point cloud data in subsequent frames. Regardless of whether the data is two-dimensional or three-dimensional, the representation can be chosen as a grid or voxel, offering more options for smooth map reconstruction \cite{rusinkiewicz2001efficient}. 

However, although LiDAR schemes boast high accuracy, they may encounter challenges in specific scenes where the density of point cloud data is insufficient for iterative matching, often resulting in map inaccuracies, such as in empty scenes or unchanging corridors. To address the limitations of individual schemes, sensor data fusion is commonly employed in the fields of autonomous driving, drones, and robotics. For example, LiDAR can be integrated with visual information or other sensors like odometers, GPS, satellite navigation systems, and inertial measurement units (IMUs). In indoor scenarios, such as storage robots or indoor service robots, two-dimensional LiDAR-based mapping methods are often employed. On the other hand, three-dimensional LiDAR-based SLAM mapping is frequently employed in the drone or autonomous driving field. Thus, this work primarily adopts a two-dimensional LiDAR-based scheme, integrating radar data with odometer and IMU data to achieve accurate robot positioning and obtain map information of the surrounding environment.

When it comes to planning and control problems, optimal control theory is often employed to find solutions. This theory aims to optimize the state of the system at each time step by utilizing quadratic programming or least square formulations to derive an optimal input continuously \cite{li2019mpc} and plan desired path \cite{mercy2017spline}, \cite{febbo2017moving}. Examples include model-based predictive control (MPC) and Timing Elastic Band (TEB) methods \cite{rosmann2017kinodynamic}. This method is capable of predicting nonlinear dynamic models or kinematic models and can handle multiple constraints. It primarily relies on the robot's kinematics or dynamics model and various constraints to formulate an optimization problem that is then solved using a solver. However, since the state of the robot changes during motion, it is necessary to continuously calculate the predicted step size for the control inputs and output quantities during the optimal control process. 

Another approach is based on velocity sampling, such as the Dynamic Window Method (DWA) \cite{seder2007dynamic}. This method also relies on the robot's chassis model to generate trajectories over a given time period at different sampling speeds. It considers various constraints, including environmental obstacles and robot chassis speed, and generates an evaluation function for different paths. The optimal path is then selected based on this evaluation function. Compared to the optimal control method, the sample-based method involves lower computational complexity and offers advantages in real-time applications. 

Indeed, both the optimal control method and the sample-based method can yield better planning and tracking outcomes. However, their computational demands can be substantial. To mitigate this challenge, practical implementations commonly employ a two-module approach for navigation: global planning and local trajectory tracking. The global planning module concentrates on generating a high-level path by considering the real-time environment. It determines the optimal route or path for the robot to follow. On the other hand, the local trajectory tracking module is responsible for precisely adhering to the generated path in real-time. This division allows for efficient management of the computational load, ensuring smooth and effective navigation in practical applications \cite{debrouwere2015optimal}. For global planning, A* and Dijkstra algorithms are developed based on the shortest path between the starting point and the target point \cite{hart1968formal}, \cite{stentz1995focused}. Solutions can also be obtained via sampling based methods, like quick search approach for random numbers \cite{lavalle2001randomized}. 

\section{Methodology}

\subsection{Kinematic Modelling}
In this work, we consider that the main environment of the chassis robot is flat ground, and the tires are hard material. In addition, in order to simplify the modeling process and adapt to the open source algorithm framework in the robot operating system, the following kinematic model is considered:

\begin{figure}[htbp]
\centerline{\includegraphics[width=0.7\linewidth]{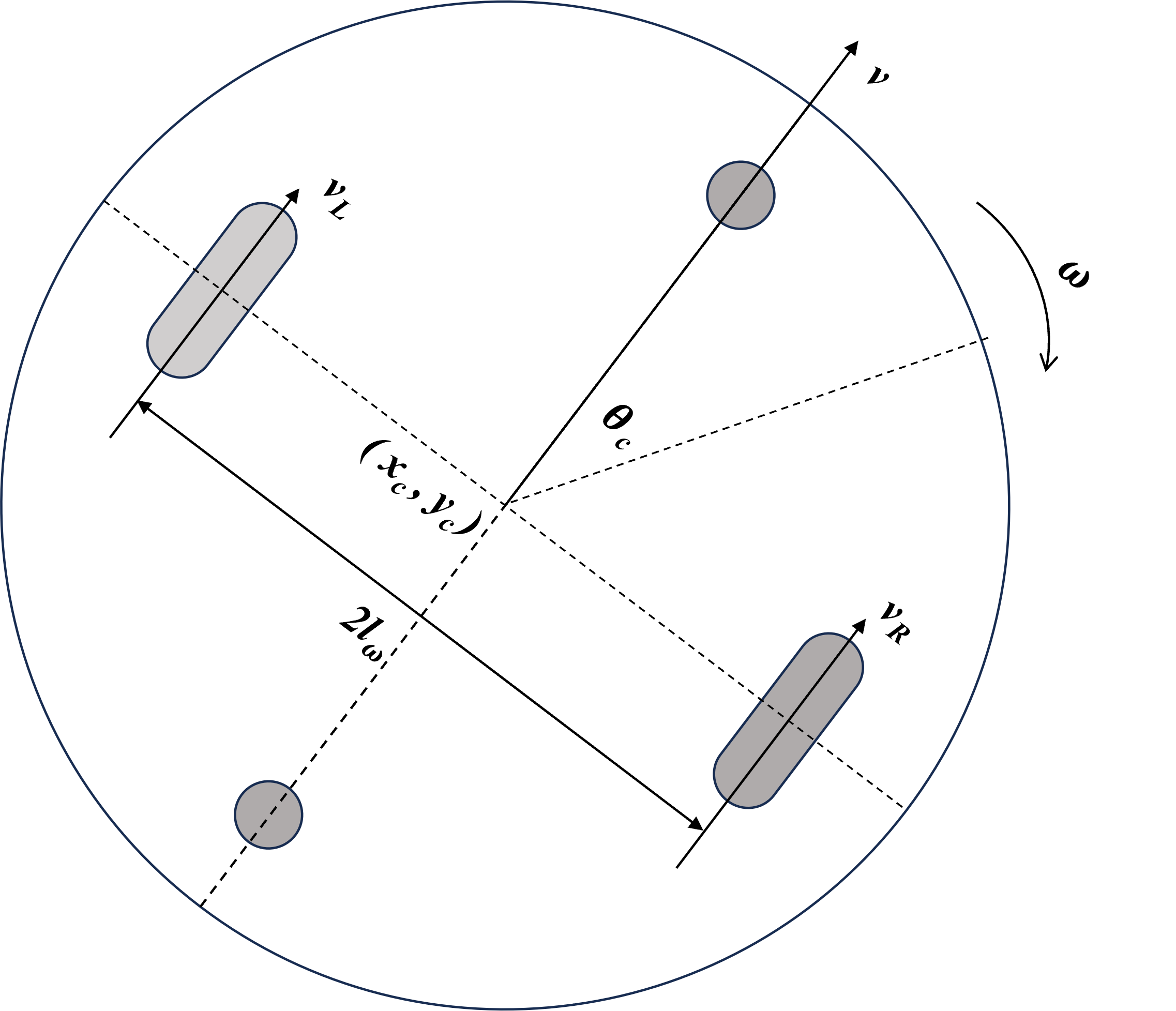}}
\caption{The mobile platform of Curzr.}
\label{fig1}
\end{figure}

As shown in Fig. \ref{fig1}, let $\begin{bmatrix}{x_c,y_c}\end{bmatrix}^T$ be the center of the chassis and $\theta_c$ be the direction of rotation of the robot. Let us denote $pos = \begin{bmatrix}x_c,y_c, \theta_c\end{bmatrix}^T$ to represent the attitude information of the chassis robot, and the distance between the two driving wheels is also defined as $2l_w$. The kinematics equation of the chassis can be expressed as follows:

\begin{equation}
\label{eq1}
\begin{cases}
\overset{\cdot}{x}_c(t)=v(t)cos(\theta_c(t)), \\
\overset{\cdot}{x}_c(t)=v(t)sin(\theta_c(t)), \\
\overset{\cdot}{\theta}_c(t)=\omega(t). \\
\end{cases}  
\end{equation}      
In system (\ref{eq1}), $v(t)$ denotes the linear velocity and $\omega(t)$ denotes the angular velocity. Meanwhile, in order to meet the dynamics characteristics of the chassis, the velocity $v_L$ of the left wheel and the velocity $v_R$ of the right wheel hold the following inequality:

\begin{equation}
\label{eq2}
\begin{cases}
\vert v^L \vert \leq \nu_{max},\\
\vert v^R \vert \leq \nu_{max}.\\
\end{cases}
\end{equation}
There is a relationship between the linear and angular velocities obtained from the kinematic model of the chassis and the left and right wheels of the chassis as follows:
\begin{equation}
\label{eq3}
\begin{cases}
v = (v^L + v^R) / 2,\\
\omega = (v^R - v^L) / 2l_{\omega}.\\
\end{cases}
\end{equation}
Based on the above setup, the linear speed assigned by the chassis to the left and right wheels of the chassis can be calculated according to the linear speed and angular speed of the chassis, so as to achieve the movement of the chassis in all directions.

Equation (\ref{eq1}) belongs to the chassis kinematics model under continuous time, but in real application, the continuous time model needs to be discretized by Euler method or Runge-Kutta method. Its discretization model is represented by:

\begin{equation}
\label{eq4}
\left[
\begin{matrix}
x_c(k+1) \\
y_c(k+1) \\
\theta_c(k+1) \\
\end{matrix}
\right] = \left[
\begin{matrix}
x_c(k) \\
y_c(k) \\
\theta_c(k) \\
\end{matrix}
\right] + \left[
\begin{matrix}
cos(\theta_c(k)) 0 \\
sin(\theta_c(k)) 0 \\
0 1 \\
\end{matrix}
\right]\left[
\begin{matrix}
v(k) \\
\omega(k) \\
\end{matrix}
\right]\Delta T.
\end{equation}

\begin{figure}[htbp]
\centerline{\includegraphics[width=1.0\linewidth]{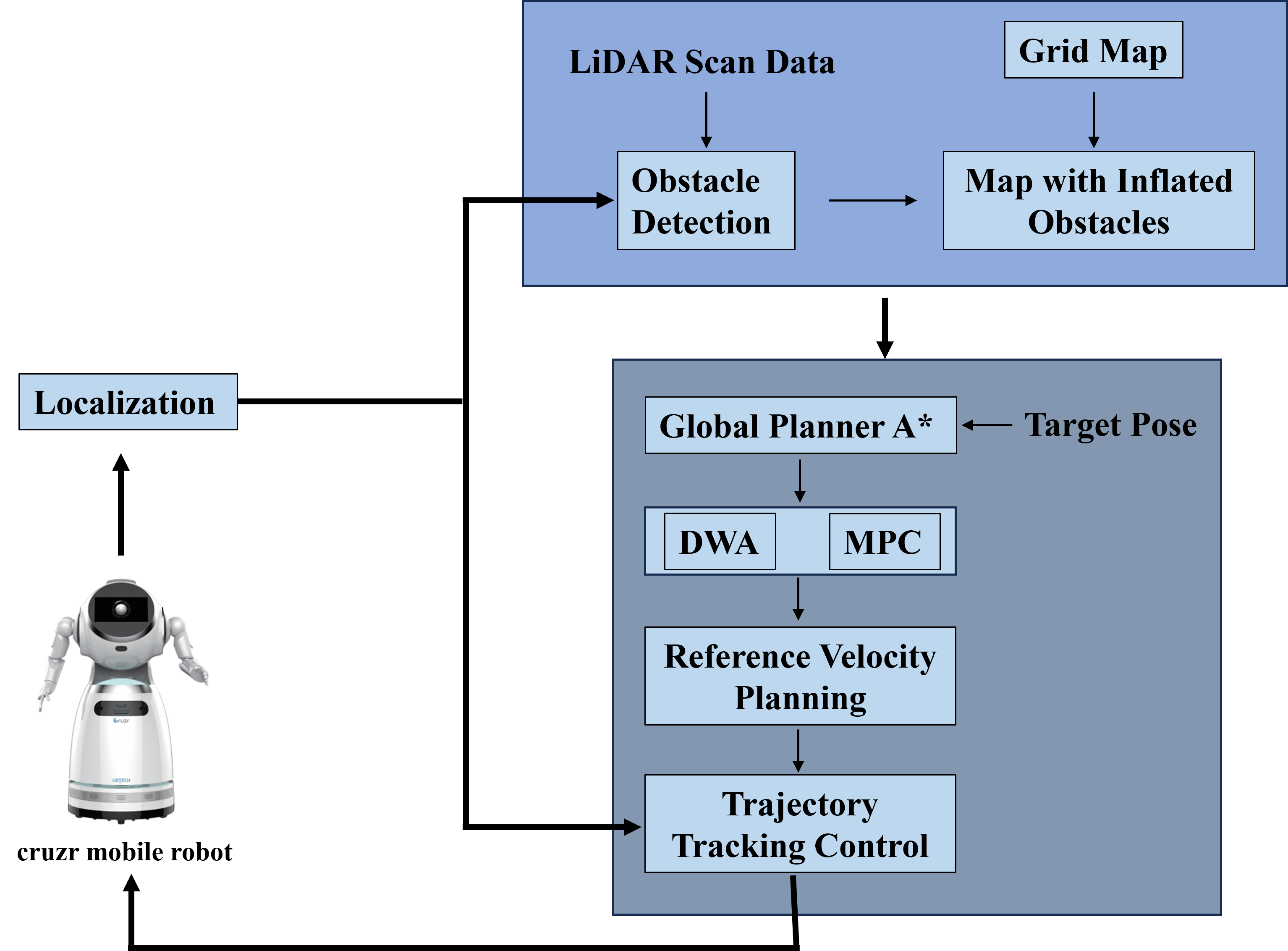}}
\caption{The navigation framework of Cruzr.}
\label{fig}
\end{figure}

\subsection{Navigation Framework}
The navigation of a robot typically consists of three main components: obstacle map construction, global planning, and local planning. Firstly, the obstacle map is constructed using the slam-gmapping algorithm. This algorithm utilizes adaptive Monte Carlo Localization (AMCL) and particle filters to estimate the robot's current pose. It combines radar scanning data in real-time to generate a 2D map of the environment. The constructed map is then rasterized, and the obstacles within it are expanded to ensure safe navigation. The global planner utilizes the A* algorithm to generate a global path. This path is based on a global cost map that is updated with a low frequency. The global planner focuses on finding an optimal path to the goal while considering the overall cost of traversal.

On the other hand, the local planner employs trajectory tracking algorithms such as Dynamic Window Approach (DWA) or Model Predictive Control (MPC). These algorithms operate based on local cost maps that are updated in real-time. The local cost map is generated dynamically according to the real-time environment, enabling the local planner to perform dynamic obstacle avoidance. The fast update frequency of the local cost map allows for real-time adaptation and ensures safe navigation in the presence of moving obstacles. Overall, the combination of obstacle map construction, global planning, and local planning enables the robot to navigate effectively and avoid obstacles in real-time scenarios.

\subsection{SLAM-Gmapping}
Gmapping uses LiDAR sensor data and motion information to locate and map simultaneously. It estimates the robot's attitude and the probability distribution of the map using the Adaptive Monte Carlo Localization algorithm and a raster map representation. By comparing the measured data with the predicted data based on the robot's motion model, Gmapping updates the map and gradually builds an accurate representation while tracking the robot's real-time position on the map. The pseudo-code is shown below:


\begin{algorithm}
    \caption{Gmapping}
    \hspace*{\algorithmicindent} \textbf{Input:} AMCL-Filter, Laser-scan \\
    \hspace*{\algorithmicindent} \textbf{Output:} Map
        \begin{algorithmic}[1]
            \State Initialize map, amcl filter ${nums}$, laser scan ${arr}$.
            \State { }while(${arr}$) do
            \State { }{ }{ }Update the kinematics model of amcl filter.
            \State { }{ }{ }for ${i}$ = 0 to ${nums}$
            \State { }{ }{ }{ }{ }Update the map using particle pose.
            \State { }{ }{ }{ }{ }Map updates for all particles.
            \State return map
        \end{algorithmic}
 \end{algorithm}

 During the actual use of the Cruzr, there is an issue with tire vibration during rotation. This vibration affects the accuracy of the mileage meter integration, causing the positioning information of the robot to drift. As a result, the slam (simultaneous localization and mapping) algorithm, specifically gmapping, is unable to build the map correctly. To address this issue, the angular speedometer and odometer were utilized to observe the linearized system model of the odometer. The observed data from these sensors were then fused and filtered to enhance the stability of external disturbances. The measurement noise covariance matrix of the Extended Kalman Filter (EKF) can be directly obtained from the sensor.

  \begin{algorithm}
    \caption{EKF}
    \hspace*{\algorithmicindent} \textbf{Input:} odom data ${o = [pos]}$ \\
    \hspace*{\algorithmicindent} \textbf{Output:} odom data ${o_{ekf} = [pos]}$
        \begin{algorithmic}[1]
            \State Initialize sensor's model and measurement model.
            \State Initialize system state and covariance matrix ${Q},{R}$.
            \State { }while(${o = [pos]}$) do
            \State { }{ }{ }State prediction.
            \State { }{ }{ }State update.
            \State return ${o_{ekf} = [pos]}$
        \end{algorithmic}
 \end{algorithm}

\subsection{Comparison of DWA and MPC Local Planner}
The pseudo-code below describes the model predictive control (MPC) process. It involves setting a time step and using a discretized robot kinematics model to calculate the future trajectory of movement. The calculation takes into account various hard and soft constraints, and aims to find the optimal control input value using quadratic programming or least squares form.

 \begin{algorithm}
    \caption{MPC Local Planner}
    \hspace*{\algorithmicindent} \textbf{Input:} global path ${r = [x,y]}$, map \\
    \hspace*{\algorithmicindent} \textbf{Output:} local path ${{r}^{*} = [x,y,\theta]}$
        \begin{algorithmic}[1]
            \State Initialize kinematics model and controller's param.
            \State { }while(${r = [x,y]}$, ${{r}_{now}}$) do
            \State { }{ }{ }Calculate the optimal control input based on kinematic model.
            \State { }{ }{ }Pub the optimal input.
            \State return ${{r}^{*}}$
        \end{algorithmic}
 \end{algorithm}

DWA is a method used in local planning for generating and evaluating different trajectories for a robot. It operates by sampling different speeds within the motion space and generating a set of trajectories over a specified time period. These trajectories are then evaluated using a scoring function, and the trajectory with the highest score is selected as the local planning path.

  \begin{algorithm}
    \caption{DWA Local Planner}
    \hspace*{\algorithmicindent} \textbf{Input:} global path ${r = [x,y]}$, map \\
    \hspace*{\algorithmicindent} \textbf{Output:} local path ${{r}^{*} = [x,y,\theta]}$
        \begin{algorithmic}[1]
            \State Initialize kinematics model and controller's param.
            \State { }while(${r = [x,y]}$, ${{r}_{now}}$) do
            \State { }{ }{ }get sampling speed ${{s} = [\theta, {v}]}$
            \State { }{ }{ }for ${i}$ = 0 to ${s}_{nums}$
            \State { }{ }{ }{ }{ }Generate the traj based on kinematic model.
            \State { }{ }{ }{ }{ }Score for the generated traj.
            \State { }{ }{ }{ }{ }Choose the highest score from the traj.
            \State { }{ }{ }{ }{ }Pub the highest score traj.
            \State return ${{r}^{*}}$
        \end{algorithmic}
 \end{algorithm}

\subsection{Hidden Markov Model for Text-to-Speech Technology}
Hidden Markov Models (HMMs) belong to a type of dynamic Bayesian models and have their own unique structure, making them suitable for various applications (Kayte, 2015). HMMs are used to model systems that have a Markov process with hidden states. In its simplest form, an HMM model consists of a state variable (S), an observed variable (O), and transitions between states with associated probabilities. In other words, HMMs are graph models that can effectively predict hidden variables based on observable variables. For example, an HMM can be used to predict the weather in a location based on the clothing people are wearing. One of the key advantages of using HMMs is the Markov assumption, which states that future events are independent of past events and only depend on the current state. This means that if we know the current state, we can predict the future state without needing additional training data. HMMs are commonly used in fields like speech recognition, particularly in intensive chemistry programs such as speech-to-text (STT) and text-to-speech (TTS) systems \cite{kayte15ijca}.

HMMs are also used to train Text-to-Speech (TTS) transformation models in order to achieve effective results. Fig. \ref{chatbot} shows the basic steps involved in TTS conversion using HMM. The steps can be summarized as follows:
\begin{itemize}
    \item Text information storage: The system receives text input from the user during runtime and stores it.
    \item Preprocessing of stored messages: This step involves removing any irregular or noisy data from the text input. Additionally, the text is divided into overlapping text frames.
    \item HMM Training: HMM is a statistical model used for text recognition. In this step, the HMM model is trained to simulate an unknown system based on observed output sequences. This includes tasks such as pattern recognition, mapping words to their sounds, and creating pattern representations of features extracted from text classes using one or more test patterns associated with speech sounds of the same class \cite{trivedi18jce}.
    \item Voice Message Recognition: Finally, speech is generated as an output and assembled, resulting in a speech output.
\end{itemize}

\subsection{Development of Intelligent Chatbot Application}
In order to apply intelligent voice technology to real-life situations and provide convenience to users, we have designed and developed a voice Chatbot APP using Android Studio, an application development software. The APP includes a basic login and registration interface for users to create their own accounts. The main function of the app is to allow users to send text messages directly through the intelligent voice chat feature. Users can also send voice messages by recording their voice using the device's microphone. When the device receives a voice message, the relevant code converts the voice into text using STT technology and displays it on the interface.

The system then analyzes and understands the user's questions using the API provided by a technology company and Natural Language Processing (NLP) technology. Based on this analysis, the system generates a corresponding reply. To provide the reply to the user, TTS technology is used to convert the text message into voice information. The reply is then presented to the user through the interface button in the form of voice. Additionally, an additional function has been added to the app. By clicking on the voice bar of the system's reply in the user interface, the page displays the text form of the system's reply message. This provides users with different options in various situations.

 \begin{figure}[htbp]
\centerline{\includegraphics[width=1.0\linewidth]{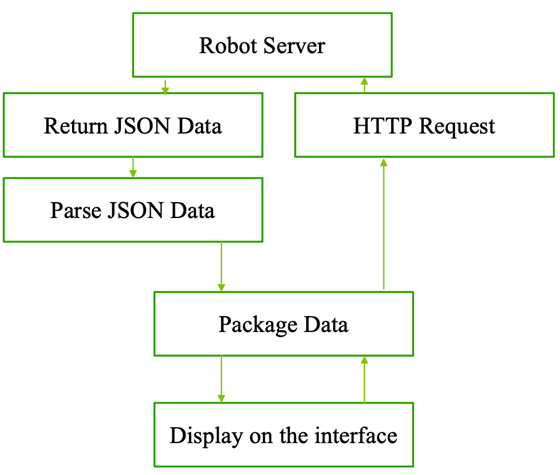}}
\caption{Structure and dataflow of Chatbot.}
\label{Chatbot}
\end{figure}

The modeling process of the intelligent chatbot can be described in the following steps:
\begin{itemize}
\item Step 1: Create a data adapter called ChatAdapter. This adapter is responsible for returning the data type and defining the arrangement style of each child in the list. It is used to adapt the data for the ListView control.

\item Step 2: Check if the user's message is not empty. If it is not empty, encapsulate the message data.

\item Step 3: Store the user's message locally as a historical message.

\item Step 4: Check if the variable "isReqApi" is true. If it is true, call the function getResponse using the getFromMsg method to send the user's question to the API for a response.

\item Step 5: Once the response is received from the API, parse the JSON dataset.

\item Step 6: Package and display the parsed JSON data on the interface.
\end{itemize}

To summarize, the modeling process involves creating a data adapter, encapsulating and storing the user's message, sending the user's question to the API, receiving and parsing the JSON response, and displaying the parsed data on the interface.

\section{Depolymnet on a Humanoid Robot: Cruzr}
\subsection{Automatic Navigation Comparison between MPC and DWC on Cruzr}
To compare the effects of two planning algorithms in real scenarios, we selected two situations. The first scenario involves obstacle avoidance, where the robot encounters multiple obstacles. The effects can be observed in the demo video accompanying the report. It is evident that the paths generated by the MPC local planner are smoother and more effective in following global paths.

 \begin{figure}[htbp]
\centerline{\includegraphics[width=1.0\linewidth]{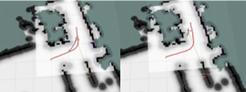}}
\caption{Trajectories generated by MPC (left) and DWA (right).}
\label{path}
\label{traj}
\end{figure}

The second scenario involves a corner where the DWA algorithm is unable to generate a local path to navigate around it. In contrast, the MPC algorithm is capable of passing through the corner, as depicted in Fig. \ref{traj}. As an optimization method, MPC can take into account dynamic constraints, resulting in more precise trajectory tracking. On the other hand, DWA sacrifices some accuracy by prioritizing the optimal speed sampling method.

\begin{figure}[htbp]
\centerline{\includegraphics[width=0.7\linewidth]{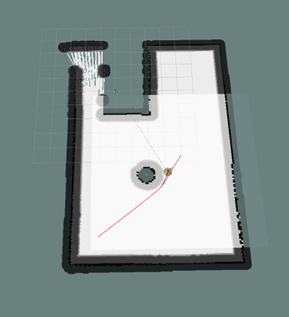}}
\caption{The case of the chassis avoiding obstacles in the simulator.}
\label{obstacle}
\end{figure}

\begin{figure}[htbp]
\centerline{\includegraphics[width=1.0\linewidth]{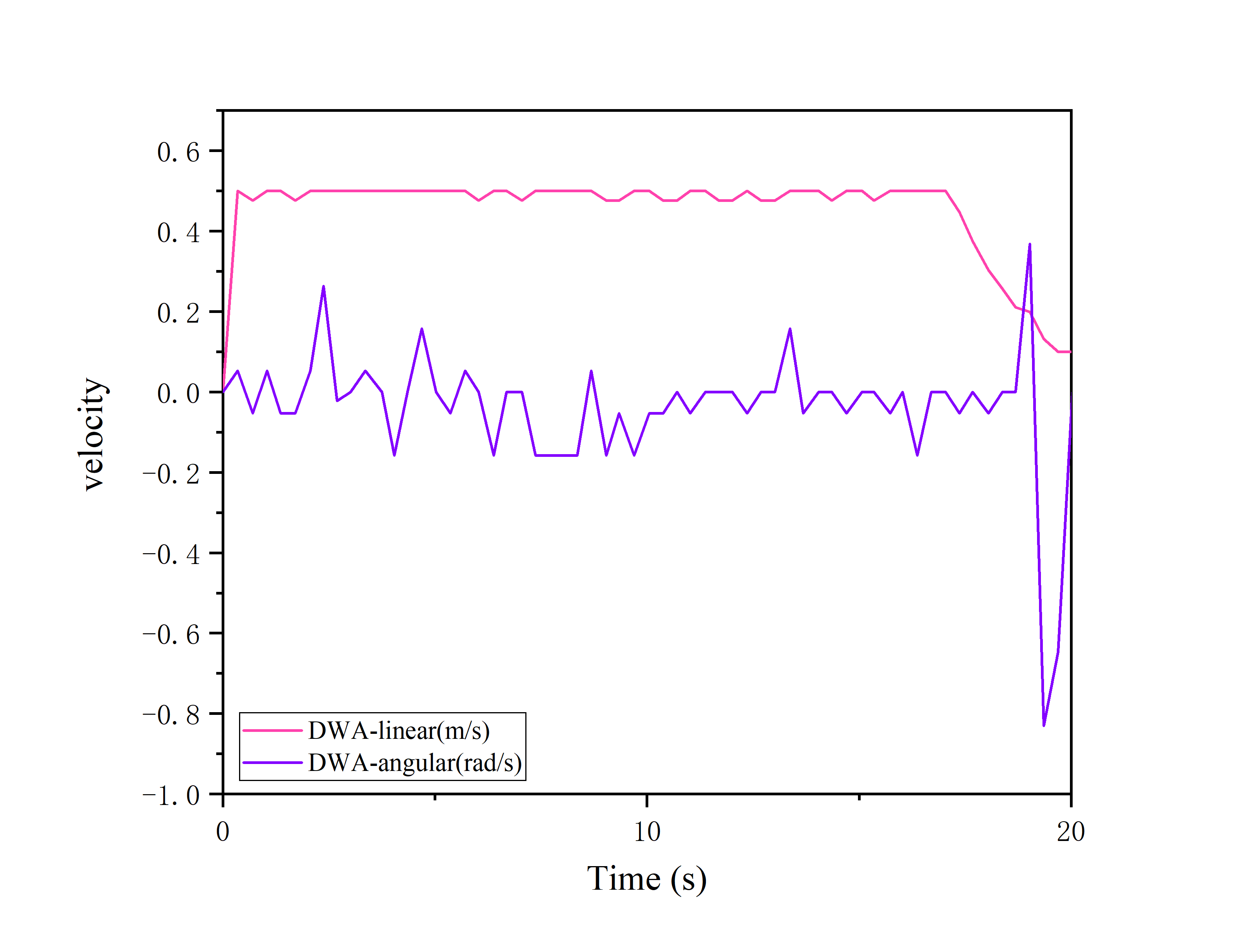}}
\centerline{\includegraphics[width=1.0\linewidth]{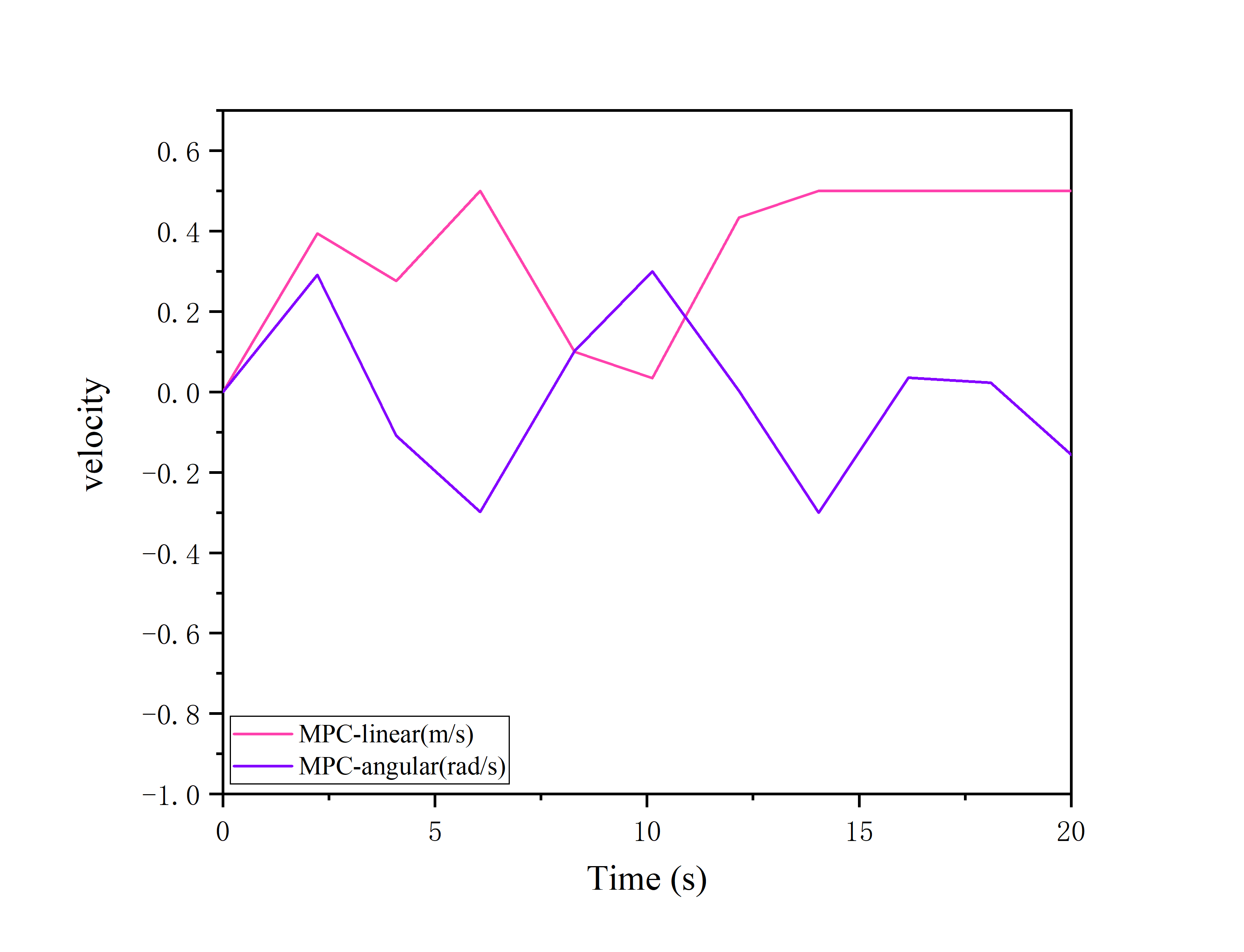}}
\caption{Speed output value of DWA and MPC.}
\label{speedoutput}
\label{speed}
\end{figure}

To demonstrate the robustness of the MPC algorithm, I conducted a simulation shown in Fig. \ref{obstacle}. In this scenario, an obstacle is placed in the path of a chassis robot moving in a straight line. Within a motion time of 20 seconds, the linear and angular velocities generated by both the MPC and DWA algorithms can be observed in Fig. \ref{speed}. Both algorithms start at zero velocity and reach the same target point. The chart clearly indicates that the control output speed under the MPC algorithm exhibits significantly fewer fluctuations compared to DWA. Additionally, the angular speed under DWA occasionally drops, reaching a minimum of -0.8 rad/s. On the other hand, the linear and angular velocities generated by MPC remain stable within a fixed range, demonstrating its robustness.

\subsection{Development of Chatbot Application on Cruzr}
\begin{figure}[htbp]
\centerline{\includegraphics[width=0.95\linewidth]{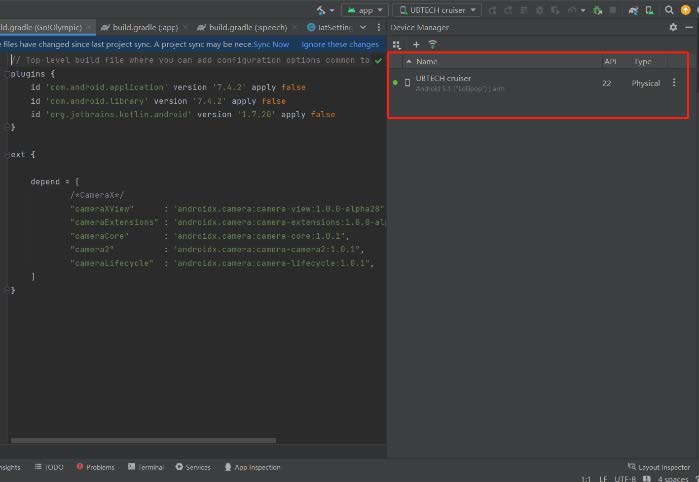}}
\caption{Cruzr robot connection to a local desktop computer.}
\label{connection}
\end{figure}

To install the developed chatbot app on the Cruzr robot, follow these steps: 1) Connect the Cruzr robot to a private network. 2) Find the IP address of the Cruzr robot through the system settings. 3) Ensure that a local desktop is connected to the same private network and is in the same network segment. 4) Use the ADB command (Cruzr internal command) to install the chatbot app on the Cruzr robot. Please refer to Fig. \ref{connection} for a visual representation of the internet connection between the Cruzr robot and the local desktop.

\begin{figure}[htbp]
\centerline{\includegraphics[width=0.7\linewidth]{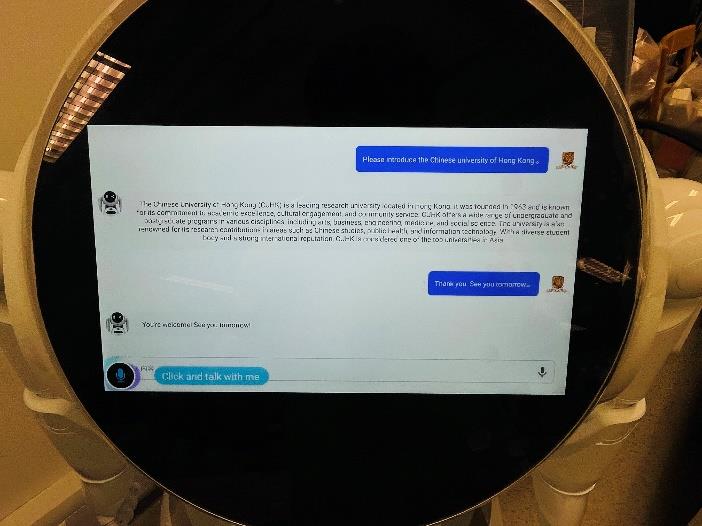}}
\caption{Chatbot testing on Cruzr.}
\label{chatbot}
\end{figure}

It was tested that Cruzr could succesfully recognize the vocal speeches in English (correctness rate is 96.7\%), and the responses from the Chatbot's performance was also with a high BLEU score (BiLingual Evaluation Understudy), which is 0.72 out of 1 in 3000 dialog testing. The interface of Chatbot is displayed in Fig. \ref{chatbot}. It validates that the developed chatbot on the Cruzr robot is capable to engage in both speech and text communication and question-and-answer interactions with human.

\section{Discussion}
The Model Predictive Control (MPC) and Dynamic Window Approach (DWA) are two distinct local path planning algorithms, each with its own advantages and disadvantages depending on the application scenario and performance indicators.

MPC is an optimization-based method that predicts future states and control inputs to select the optimal control strategy. It excels in trajectory tracking, motion smoothness, and considers dynamic constraints and target optimization. However, due to its complex optimization process, MPC requires sufficient computing resources. It performs exceptionally well in scenarios that demand precise trajectory tracking and involve complex dynamic constraints.

On the other hand, DWA is a simpler and real-time local path planning algorithm. It generates local paths by sampling and evaluating velocity combinations in the velocity space. DWA offers low computational complexity and real-time performance, making it suitable for scenarios that require high real-time performance. It performs effectively in quickly avoiding obstacles and adapting to dynamic environments. However, in scenarios that necessitate more accurate path tracking and consider more dynamic constraints, DWA may not perform as well as MPC.

Therefore, the choice between MPC and DWA depends on the specific application scenario and performance indicators. If accurate trajectory tracking and consideration of complex dynamic constraints are crucial, and computational resources are sufficient, MPC is more suitable. On the other hand, if real-time requirements are high, and quick obstacle avoidance and adaptability to dynamic environments are essential, DWA is more appropriate. Ultimately, selecting the algorithm that aligns with the specific application requirements is the best approach.

\section{Conclusion}
Mobile robots have significant potential and are expected to be widely used in the service industry. The crucial role played by automatic navigation and voice cloning cannot be overstated, as they enable functional robots to provide high-quality services. This study focuses on developing a control algorithm for the automatic navigation of Cruzr, a humanoid mobile robot manufactured by Ubtech. First, a virtual environment is constructed in the Gazebo simulation software using the Simultaneous Localization And Mapping (SLAM) technique. This allows for global path planning, which is achieved through local path tracking. The two-wheel differential chassis kinematics model is utilized to ensure that the robot chassis can autonomously avoid dynamic obstacles. Furthermore, the mapping and trajectory generation algorithms developed in the simulation environment are successfully implemented on the physical robot, Cruzr. The performance of the automatic navigation system is compared between the Dynamic Window Approach (DWA) and Model Predictive Control (MPC) algorithms. In addition to this, a mobile application is developed to enable voice cloning using a Hidden Markov Model. The proposed Chatbot is thoroughly tested and subsequently deployed on the Cruzr robot. This allows for enhanced interaction and communication capabilities, further improving the overall service quality provided by the robot. Future efforts will be devoted to barrier functions and robust synthesis by using Semi-definite Programming \cite{wang18ACC,han19tac,han2014tcas1,han12tii}.

\section*{Acknowledgment}
We would like to express our gratitude for the financial support provided by the Teaching Development and Language Enhancement Grant (4170989) from the Hong Kong SAR, China. This support has been instrumental in the successful completion of this study.

Additionally, we would like to acknowledge and appreciate the valuable inputs received from Prof. Yunhui Liu at The Chinese University of Hong Kong. His expertise and guidance have greatly contributed to the quality of this research.

\bibliographystyle{IEEEtran}
\bibliography{reference}

\begin{thebibliography}{10}
\providecommand{\url}[1]{#1}
\csname url@samestyle\endcsname
\providecommand{\newblock}{\relax}
\providecommand{\bibinfo}[2]{#2}
\providecommand{\BIBentrySTDinterwordspacing}{\spaceskip=0pt\relax}
\providecommand{\BIBentryALTinterwordstretchfactor}{4}
\providecommand{\BIBentryALTinterwordspacing}{\spaceskip=\fontdimen2\font plus
\BIBentryALTinterwordstretchfactor\fontdimen3\font minus
  \fontdimen4\font\relax}
\providecommand{\BIBforeignlanguage}[2]{{%
\expandafter\ifx\csname l@#1\endcsname\relax
\typeout{** WARNING: IEEEtran.bst: No hyphenation pattern has been}%
\typeout{** loaded for the language `#1'. Using the pattern for}%
\typeout{** the default language instead.}%
\else
\language=\csname l@#1\endcsname
\fi
#2}}
\providecommand{\BIBdecl}{\relax}
\BIBdecl

\bibitem{Taketomi2017Visual}
T.~Taketomi, H.~Uchiyama, and S.~Ikeda, ``Visual slam algorithms: a survey from
  2010 to 2016,'' \emph{IPSJ Transactions on Computer Vision and Applications},
  vol.~9, no.~16, 2017.

\bibitem{rusinkiewicz2001efficient}
S.~Rusinkiewicz and M.~Levoy, ``Efficient variants of the {ICP} algorithm,'' in
  \emph{Proceedings of IEEE International Conference on 3-D Digital Imaging and
  Modeling}, 2001, pp. 145--152.

\bibitem{li2019mpc}
J.~Li, M.~Ran, H.~Wang, and L.~Xie, ``{MPC}-based unified trajectory planning
  and tracking control approach for automated guided vehicles,'' in
  \emph{Proceedings of IEEE International Conference on Control and
  Automation}, 2019, pp. 1--6.

\bibitem{mercy2017spline}
T.~Mercy, R.~Van~Parys, and G.~Pipeleers, ``Spline-based motion planning for
  autonomous guided vehicles in a dynamic environment,'' \emph{IEEE
  Transactions on Control Systems Technology}, 2017.

\bibitem{febbo2017moving}
H.~Febbo, J.~Liu, P.~Jayakumar, J.~L. Stein, and T.~Ersal, ``Moving obstacle
  avoidance for large, high-speed autonomous ground vehicles,'' in
  \emph{Proceedings of IEEE American Control Conference}, 2017, pp. 5568--5573.

\bibitem{rosmann2017kinodynamic}
C.~R{"o}smann, F.~Hoffmann, and T.~Bertram, ``Kinodynamic trajectory
  optimization and control for car-like robots,'' in \emph{Proceedings of
  IEEE/RSJ International Conference on Intelligent Robots and Systems (IROS)},
  2017, pp. 5681--5686.

\bibitem{seder2007dynamic}
M.~Seder and I.~Petrovic, ``Dynamic window based approach to mobile robot
  motion control in the presence of moving obstacles,'' in \emph{Proceedings of
  IEEE International Conference on Robotics and Automation}, 2007, pp.
  1986--1991.

\bibitem{debrouwere2015optimal}
F.~Debrouwere, ``Optimal robot path following fast solution methods for
  practical non-convex applications,'' Ph.D. dissertation, KU LEUVEN, 2015.

\bibitem{hart1968formal}
P.~E. Hart, N.~J. Nilsson, and B.~Raphael, ``A formal basis for the heuristic
  determination of minimum cost paths,'' \emph{IEEE Transactions on Systems
  Science and Cybernetics}, vol.~4, no.~2, pp. 100--107, 1968.

\bibitem{stentz1995focused}
A.~Stentz \emph{et~al.}, ``The focused d*-algorithm for real-time replanning,''
  in \emph{Proceedings of International Joint Conference on Artificial
  Intelligence}, 1995, pp. 1652--1659.

\bibitem{lavalle2001randomized}
S.~M. LaValle and J.~J. Kuffner~Jr, ``Randomized kinodynamic planning,''
  \emph{The International Journal of Robotics Research}, vol.~20, no.~5, pp.
  378--400, 2001.

\bibitem{kayte15ijca}
S.~Kayte, M.~Mundada, and J.~Gujrathi, ``Hidden {M}arkov model based speech
  synthesis: A review,'' \emph{International Journal of Computer Applications},
  vol. 130, no.~3, pp. 35--39, 2015.

\bibitem{trivedi18jce}
A.~Trivedi, N.~Pant, P.~Shah, S.~Sonik, and S.~Agrawal, ``Speech to text and
  text to speech recognition systems-areview,'' \emph{IOSR Journal of Computer
  Engineering}, vol.~20, no.~2, pp. 36--43, 2018.

\bibitem{wang18ACC}
L.~Wang, D.~Han, and M.~Egerstedt, ``Permissive barrier certificates for safe
  stabilization using sum-of-squares,'' in \emph{Annual American Control
  Conference}, 2018, pp. 585--590.

\bibitem{han19tac}
D.~Han and D.~Panagou, ``Robust multi-task formation control via parametric
  {L}yapunov-like barrier functions,'' \emph{IEEE Transactions on Automatic
  Control}, vol.~64, no.~11, pp. 4439--4453, 2019.

\bibitem{han2014tcas1}
D.~Han and G.~Chesi, ``Robust synchronization via homogeneous
  parameter-dependent polynomial contraction matrix,'' \emph{IEEE Transactions
  on Circuits and Systems I: Regular Papers}, vol.~61, no.~10, pp. 2931--2940,
  2014.

\bibitem{han12tii}
D.~Han, G.~Chesi, and Y.~S. Hung, ``Robust consensus for a class of uncertain
  multi-agent dynamical systems,'' \emph{IEEE Transactions on Industrial
  Informatics}, vol.~9, no.~1, pp. 306--312, 2012.

\end{thebibliography}

\end{document}